\documentclass{article}

\usepackage{spconf,amsmath,graphicx}
\usepackage{dsfont}
\usepackage{amsfonts}
\usepackage{bm}
\usepackage{amssymb}
\usepackage{amsthm}
\usepackage{algorithm}
\usepackage[noend]{algpseudocode}
\usepackage{enumitem}
\usepackage{xcolor}

\newtheorem{prop}{Proposition}
\newtheorem{res}{Result}

\newtheorem{rem}{Remark}

\newenvironment{hproof}{%
	\proof}{\endproof}

\title{Optimal Laplacian Regularization\\ for Sparse Spectral Community Detection}
\name{Lorenzo Dall'Amico$^{\dagger}$ \qquad  Romain Couillet$^{\dagger\star}$ \qquad Nicolas Tremblay$^{\dagger}$}
\address{$^{\dagger}$ GIPSA-lab, Universit\' e Grenoble Alpes, CNRS, Grenoble INP \\
$^{\star}$L2S, CentraleSup\'elec, University of Paris Saclay}

\begin{document}
\ninept
\maketitle
\begin{abstract}
Regularization of the classical Laplacian matrices was empirically shown to improve spectral clustering in sparse networks. It was observed that small regularizations are preferable, but this point was left as a heuristic argument. In this paper we formally determine a proper regularization which is intimately related to alternative state-of-the-art spectral techniques for sparse graphs.
\end{abstract}
\begin{keywords}
Regularized Laplacian, Bethe-Hessian, spectral clustering, sparse networks, community detection
\end{keywords}
\section{Introduction}
Community detection \cite{fortunato2010community} is one of the central unsupervised learning tasks on graphs. Given a graph $\mathcal{G}(\mathcal{V},\mathcal{E})$, where $\mathcal{V}$ is the set of nodes ($|\mathcal{V}| = n$) and $\mathcal{E}$ the set of the edges, it consists in finding a label assignment -- $\hat{\ell}_i$ -- for each node $i$, in order to reconstruct the underlying community structure of the network. The community detection problem has vast applications in different fields of science \cite{barabasi2016network} and can be seen as the simplest form of clustering, \emph{i.e.} the problem of dividing objects into similarity classes.
We focus on unweighted and undirected graphs that can be represented by their adjacency matrices $A \in \{0,1\}^{n \times n}$, defined as $A_{i,j} = \mathds{1}_{(ij) \in \mathcal{E}}$, where $\mathds{1}_{x} = 1$ if the condition $x$ is verified and is zero otherwise.

A satisfactory label partition $\hat{\bm{\ell}}$ can be formulated as the solution  of an optimization problem over various cost functions, such as \emph{Min-Cut} or \emph{Ratio-Cut} \cite{von2007tutorial}. These optimization problems are NP-hard and a common way to find an approximate -- but fast -- solution, is by allowing a continuous relaxation  of these problems. Defining the degree matrix $D \in \mathbb{N}^{n\times n}$ as $D = {\rm diag}(A\bm{1}) = {\rm diag}(\bm{d})$, the eigenvectors associated to the smallest eigenvalues of the combinatorial Laplacian matrix  $L=D-A$ and to the largest eigenvalues of the matrix  $D^{-1}A$ (that we refer to as $L^{\rm rw}$) provide an approximate solution to the \emph{Ratio-Cut} and the \emph{Min-Cut} problems, respectively (Section 5 of \cite{von2007tutorial}). Retrieving communities from matrix eigenvectors is the core of \emph{spectral clustering} \cite{shi2000normalized,ng2002spectral}.\\

Although spectral algorithms are well understood in the \emph{dense} regime in which $|\mathcal{E}|$ grows faster than  $n$ (\emph{e.g.} \cite{Newman-2006,jin2015fast}), much less is known in the much more challenging \emph{sparse} regime, in which $|\mathcal{E}| = O(n)$. This regime is particularly interesting from a practical point of view, since most real networks are very sparse, and the degrees $d_i$ of each node $i$ scales as $d_i = (A\bm{1})_i \ll n$. Rigorous mathematical tools are still being developed, but some important results have already tackled the problem of community detection in sparse graphs. 

We focus here on how regularization helps spectral clustering in sparse networks. In the dense regime (Figure \ref{fig:spec}A), for a graph $\mathcal{G}$ with $k$ communities, the $k$ largest eigenvalues of $L^{\rm rw}$  are isolated and the respective eigenvectors carry the information of the low rank structure of the graph. In the sparse regime instead, (Figure \ref{fig:spec}.B) the isolated informative eigenvalues are lost in the \emph{bulk} of uninformative eigenvalues. Regularization avoids the spreading of the uninformative bulk and enables the recovery of the low rank structure of the matrix, as depicted in Figure \ref{fig:spec}.C. Among the many contributions proposing different types of regularization \cite{qin2013regularized,amini2013pseudo,joseph2013impact,lei_consistency_2015,le2018concentration}, we focus on the likely most promising one proposed by \cite{qin2013regularized} that recovers communities from the eigenvectors corresponding to the largest eigenvalues of the matrix $L_{\tau}^{\rm sym} = D_{\tau}^{-1/2}AD_{\tau}^{-1/2}$, where $D_{\tau} = D + \tau I_n$. The authors build a line of argument under the assumption that the graph is generated from the degree-corrected stochastic block model \cite{karrer2011stochastic} (DC-SBM). In the literature, the characterization of the parameter $\tau$ was never properly addressed and its assignment was left to a heuristic choice. More specifically, both in \cite{qin2013regularized} and \cite{joseph2013impact} the results provided by the authors seem to suggest a large value of $\tau$, but it is observed experimentally that smaller values of $\tau$ give better partitions. In the end, the authors in~\cite{qin2013regularized}  settle on the choice of $\tau = \frac{1}{n}\bm{1}^TD\bm{1}$, \textit{i.e.}, the average degree.

A fundamental aspect of community detection on sparse graphs generated from the DC-SBM, defined in Equation \eqref{eq:DCSBM}, is the existence of an information-theoretic threshold for community recovery \cite{decelle2011asymptotic,mossel2012stochastic,massoulie2014community,gulikers2015impossibility}: if the parameters of the generative model do not meet certain conditions, then no algorithm can assign the labels better than random guess.\\
%{\nico **sparse DC-SBM is only one model out of many sparse graph models. The existence of a threshold is NOT an aspect of ALL sparse graphs, but of some models of sparse graphs such as sparse SBM or sparse DC-SBM**},
% defined in Equation \eqref{eq:DCSBM}:
% For $k = 2$ classes, there exists a parameter $\alpha$ that controls the hardness of the detection problem, and non-trivial community recovery (\emph{i.e.} better than random assignment) is feasible if and only if $\alpha > \alpha_c$ \cite{decelle2011asymptotic,mossel2012stochastic,massoulie2014community,gulikers2015impossibility}. For more than two classes there exists a region of parameters in which community reconstruction is impossible, a region where it can be performed in exponential time and one in which the solution can be obtained in polynomial time \cite{decelle2011asymptotic}. \\

In this article we study the problem of community detection on a network generated from the sparse DC-SBM and show why a small value of $\tau$ is preferable, drawing a connection to other existing algorithms based on the Bethe-Hessian matrix \cite{saade2014spectral,dall2019revisiting}, coming from statistical physics intuitions. We further show for which value of $\tau$
%{\nico **unclear: what is the proposed regularization?**} 
the leading eigenvectors of  $L_{\tau}^{\rm rw} = D_{\tau}^{-1}A$ 
%{{\nico **$L_{\tau}^{\rm rw}$ has not been defined**}}
(and  equivalently $L_{\tau}^{\rm sym}$) allow for non-trivial  community reconstruction as soon as  theoretically possible, addressing a question not answered in \cite{qin2013regularized,amini2013pseudo,lei_consistency_2015}. The correct parametrization of $\tau$ depends on the hardness of the detection problem and, for our proposed choice of $\tau$, the matrix $L_{\tau}^{\rm rw}$ has an eigenvector, corresponding to an isolated eigenvalue whose entry $i$ only depends on the class label of node $i$ and can be used to retrieve the community labels. \\
%{\nico **following the text, I am still in the $k$ class scenario, whereas this last sentence is appropriate for $k=2$, right? (you talk about AN eigenvector associated to A isolated eigenvalue)}**{\lorenzo it is a bit annoying for us: there are $k$ values of $\tau$ and for each of them there is \textbf{one} clean eigenvector... the sentence per se is correct, but I agree it can be misleading.}

% Our result suggests that the optimal regularization should be chosen as a function of the hardness of the detection problem and we observe that in easy scenarios $L_{\tau}^{\rm rw}$ falls onto $L^{\rm rw}$. {\nico ** this contribution paragraph needs attention. State results more clearly. In the last sentence for instance, the first part is a major contribution, the last part (that $L_{\tau}^{\rm rw}$ falls onto $L^{\rm rw}$ in easy scenarios) is --almost-- a side comment.**}\\

The remainder of the article is organized as follows: in Section \ref{sec:model} we present the generative model of the graph
%{\nico **replace "network" by "graph", apart maybe from the introduction**}
and the theoretical results about the detectability transition; in Section \ref{sec:main} we give the main result together with its algorithmic implementation; Section \ref{sec:conclusion} closes the article.

% {\nico **Globally, you need to work on your introductions to make them clearer and more to-the-point. For instance, I do not think it is clear to the reader here that you propose a parametrization that is motivated and analyzed in the DC-SBM scenario which is ONE --out of MANY-- possible models of sparse graphs with community structure. Imagine yourself knowing only Von Luxburg and reading this introduction**}

\medskip

\noindent {\bf Notations.} Matrices are indicated with capital ($M$), vectors with bold (${\bm v}_p$), scalar and vector elements with standard ($v_{p,i}$) letters. We denote by $s_i^{\uparrow}(M)$ the $i$-th smallest eigenvalue of a Hermitian matrix $M$ and by $s_i^{\downarrow}(M)$ the $i$-th largest. $s_i(M)$ indicates a generic eigenvalue of $M$.
	%{\nico ** I personally find this notation cumbersome. Can't we just order the eigenvalues $\{s_i\}$ of a matrix $M$ as $s_1\leq s_2\leq\ldots\leq s_n$ and talk about $s_i$ as the $i$-th smallest and $s_{n-i}$ as the $i$-th largest?**}{\lorenzo ** The point is that $s_i(\cdot)$ is a function of the matrix, so you don't need a different notation each time, but is clear who is who. For the arrows is to distinguish the case in which you know the position of a given eigenvalue to the case in which you don't.}
The notation $M\bm{x} \approx 0$ indicates that, for all large $n$ with high probability, $\bm x$ is an approximate eigenvector of $M$ with eigenvalue $o_n(1)$.
%{\nico **this statement is super vague! but probably there is no need to make it clearer**}

\section{Model}
\label{sec:model}

Consider  a graph $\mathcal{G}(\mathcal{V},\mathcal{E})$ with $|\mathcal{V}| = n \gg 1$. We consider the DC-SBM \cite{karrer2011stochastic} as a generative model for the $k$-class graph $\mathcal{G}$. Letting $\bm{\ell} \in \{1,\cdots,k\}^n$ be the vector of the  true labels of a $k$-class network, the DC-SBM generates edges independently according to 
\begin{align}
\mathbb{P}(A_{ij} = 1|\ell_i,\ell_j,\theta_i,\theta_j) = \theta_i\theta_j\frac{C_{\ell_i,\ell_j}}{n}.
\label{eq:DCSBM}
\end{align}
The vector $\bm{\theta}$ allows for any degree distribution on the graph and satisfies $\frac{1}{n}\sum_{i \in \mathcal{V}} \theta_i = 1$ and $\frac{1}{n}\sum_{i \in \mathcal{V}} \theta_i^2 = \Phi = O_n(1)$. The matrix $C \in \mathbb{R}^{k \times k}$ is the class affinity matrix. Letting $\Pi = {\rm diag}(\bm{\pi}) \in \mathbb{R}^{k \times k}$, where $\pi_{p}$ is the fraction of nodes having label $p$, we assume that $C\Pi \bm{1} = c\bm{1}$, where it is straightforward to check that $c = \mathbb{E}[\frac{1}{n}\bm{1}^TD\bm{1}] = O_n(1)$ is the expected average degree, while, denoting with $d_i = D_{ii}$ the degree of node $i$, $\mathbb{E}[d_i] = c\theta_i$. This is a standard assumption \cite{decelle2011asymptotic,krzakala2013spectral,bordenave2015non,zdeborova2016statistical} that means that the expected degree of each node does not depend on its community, hence that the degree distribution does not contain any class information. 
%Finally, the term $\frac{1}{n}$ in Equation \eqref{eq:DCSBM} sets the problem in the sparse regime.

Considering the model of Equation \eqref{eq:DCSBM} for $k=2$ and $\bm{\pi} \propto \bm{1}_2$, we denote $C_{\ell_i,\ell_j} = c_{\rm in}$ if $\ell_i = \ell_j$ and $C_{\ell_i,\ell_j} = c_{\rm out}$ otherwise. As shown in \cite{decelle2011asymptotic,mossel2012stochastic,massoulie2014community}, non-trivial community reconstruction is theoretically feasible if and only if
\begin{equation}
\alpha \equiv \frac{c_{\rm in} - c_{\rm out}}{\sqrt{c}} > \frac{2}{\sqrt{\Phi}} \equiv \alpha_c.
\label{eq:th}
\end{equation}
The parameter $\alpha$ regulates the hardness of the detection problem: for large $\alpha$ we have easy recovery, for $\alpha \leq \alpha_c$ the problem has asymptotically no solution.
When $k > 2$ we distinguish two transitions: one from impossible to hard detection (the solution can be obtained in exponential time) and one from hard to easy detection (the solution can be obtained in polynomial time) \cite{decelle2011asymptotic}. 
% It is conjectured \cite{krzakala2013spectral} that the problem is in the easy detection regime if $s_1^{\uparrow}(C\Pi) > \sqrt{\frac{c}{\Phi}}$. {\nico are we sure we need this information?} {\lorenzo probably we don't...}

\section{Main result}
\label{sec:main}

In this section we study the relation between the matrix $L_{\tau}^{\rm rw} = D_{\tau}^{-1}A$ and the Bethe-Hessian matrix \cite{saade2014spectral}, defined as
\begin{equation}
H_r = (r^2-1)I_n + D - rA, \quad r \in \mathbb{R}.
\label{eq:Hr}
\end{equation}
We exploit some important results concerning the  spectrum of $H_r$ to better understand why regularization helps in sparse networks.

\subsection{Relation between $L_{\tau}^{\rm rw}$ and the Bethe-Hessian matrix}

In \cite{saade2014spectral} it was shown that the Bethe-Hessian matrix can be efficiently used to reconstruct communities in sparse graphs. 
% The Bethe-Hessian matrix is defined as  
% \begin{equation}
% H_r = (r^2-1)I_n + D - rA, \quad r \in \mathbb{R}.
% \label{eq:Hr}
% \end{equation}
This matrix comes from the strong connection existing between the problem of community detection and statistical physics. The authors of \cite{saade2014spectral} originally proposed to perform spectral clustering with the $k$ smallest eigenvectors of $H_r$ for $r = \sqrt{c\Phi}$. For this choice of $r$, if the problem is in the easy (polynomial) detectable regime, then only the $k$ smallest eigenvalues of $H_r$ are negative, while $s_{k+1}^{\uparrow}(H_r) \approx 0$.
% {\nico **too many typos in this sentence to be understandable**}. 
In \cite{dall2019community} we refined this approach in a two-class setting, showing that there exists a parametrization -- depending on the clustering difficulty
% {\nico **I am not sure the reader understands what you mean by hardness: it is the first the term appears in the text. You should for example name $\alpha$ as the hardness parameter in Sec.2**}
-- that leads  to better partitions under a generic degree distribution and, at the same time, provides non-trivial clustering as soon as theoretically possible. In \cite{dall2019revisiting} we extended our reasoning for more than two classes and studied the shape of the informative eigenvectors. We here recall our main findings.

% \begin{res}[from \cite{dall2019revisiting}]
% The equation $s_p^{\uparrow}(H_r) \approx 0$ has $k+1$ solutions in $r \in (1, \sqrt{c\Phi})$:
% \begin{equation}
% r = \zeta_p = \frac{c}{s_p^{\downarrow}(C\Pi)} \quad {\rm for}~ 1\leq p \leq k
% \label{eq:zeta}
% \end{equation}
% and $r = \sqrt{c\Phi}$ for $p = k+1$. The eigenvector solution to $H_{\zeta_p}\bm{x}_p \approx 0$ brings information about  communities and its entries depends -- in expectation -- only on the class labels: $\mathbb{E}[x_{p,i}] = v_{p,i}$, where $C\Pi\bm{v}_p = s_p^{\downarrow}(C\Pi)\bm{v}_p$.
% \end{res}

% {{\nico **once again, the formulation could be clearer. For instance your first sentence "The equation $s_p^{\uparrow}(H_r) \approx 0$ has $k+1$ solutions in $r \in (1, \sqrt{c\Phi})$" is not understandable as is --even though you explain what you mean afterwards. Why not be clear from the start? For instance, I suggest the following formulation. I know these comments may seem as only details: they are not though. The clearer you are, the more people will read you. Also, the more enjoyable is the reading, the more efforts people will make to actually try to understand your results.**}}
\begin{res}[from \cite{dall2019revisiting}]
Let $p$ be an integer between $1$ and $k$. The equation $s_p^{\uparrow}(H_r) \approx 0$ is verified for
\begin{equation}
r = \zeta_p = \frac{c}{s_p^{\downarrow}(C\Pi)} \quad \in (1, \sqrt{c\Phi}).
\label{eq:zeta}
\end{equation}
For $p=k+1$, the solution is $r = \sqrt{c\Phi}$. The eigenvector solution to $H_{\zeta_p}\bm{x}_p \approx 0$  has a non-trivial alignment to the community label vector and
%the expectation of its entries depends solely on the class labels: 
$\mathbb{E}[x_{p,i}] = v_{p,i}$, where $C\Pi\bm{v}_p = s_p^{\downarrow}(C\Pi)\bm{v}_p$.
\end{res}

 \begin{figure}[t!]
	\centering
	\includegraphics[width = \columnwidth]{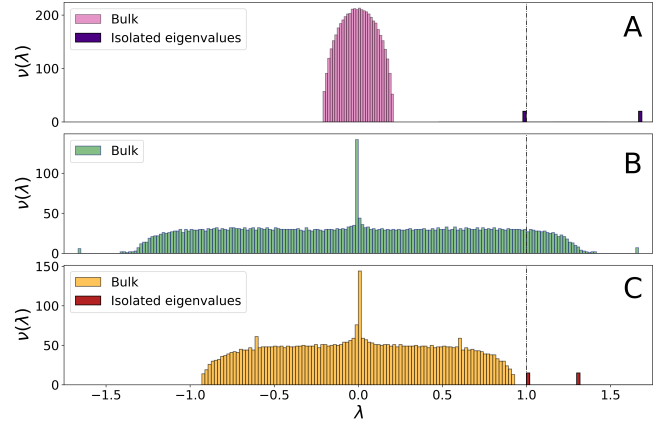}
	\caption{In all three figures $r = \zeta_2 = \frac{c_{\rm in} + c_{\rm out}}{c_{\rm in} - c_{\rm out}}$, $n = 5000$, $\theta_i \sim [\mathcal{U}(3,7)]^3$, $k = 2$. {\bf A:} Spectrum of $rD^{-1}A$ in the dense regime ($c_{\rm in}/n = 0.08$, $c_{\rm out}/n = 0.02$). {\bf B:} Spectrum of $rD^{-1}A$ in the sparse regime ($c_{\rm in} = 8$, $c_{\rm out} = 2$). {\bf C:} Spectrum of $rD^{-1}_{r^2-1}A$ in the sparse regime ($c_{\rm in} = 8$, $c_{\rm out} = 2$).}
	\label{fig:spec}
\end{figure}

This result implies that, for $r = \zeta_p$, the $p$-th smallest eigenvalue of $H_{\zeta_p}$ is close to zero and the corresponding eigenvector $\bm{x}_p$ is a noisy version of the corresponding eigenvector $\bm{v}_p \in \mathbb{R}^k$ related to the  $p$-th largest eigenvalue of $C\Pi$. Importantly, $\mathbb{E}[x_{p,i}]$ does not depend on $d_i$, hence it is suited to reconstruct communities regardless of the degree distribution. Since the eigenvectors of $C\Pi$ constitute a sufficient basis to identify the classes, the vectors $\bm{x}_p,~{\rm for}~ 1\leq p \leq k$, can be exploited to recover the community labels by stacking them in the columns of a matrix  $X \in \mathbb{R}^{n \times k}$ and running the \emph{k-means} algorithm on the rows of $X$.

\begin{rem}
	Having eigenvectors whose entries are, in expectation, independent of the degree distribution is of fundamental importance in the \emph{k-means} step. If this were not the case, then the class information would be affected by the uninformative degree distribution, compromising the performance of the algorithm, as shown in \cite{dall2019revisiting}.
	\label{rem:1}
\end{rem}

\subsection{Improved regularization for the regularized random walk Laplacian $L_{\tau}^{\rm rw}$}
\label{sec:main.proof}

\begin{figure}
    \centering
    \includegraphics[width = \columnwidth]{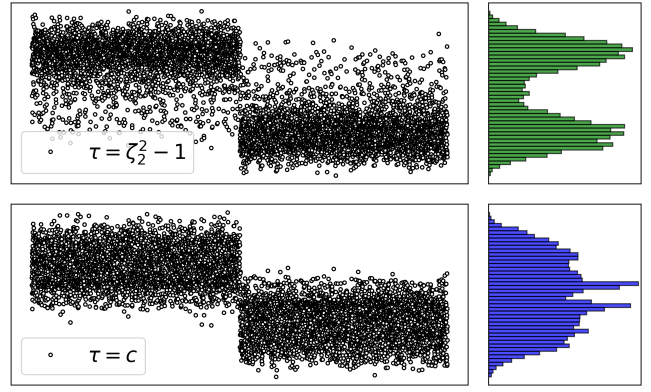}
    \caption{Eigenvector of second largest eigenvalue of $L_{\zeta_2^2-1}^{\rm rw}$ (top) and $L_{c}^{\rm rw}$ (bottom) with histogram of the densities of the entries of the eigenvector (right). For this simulation, $n = 15000$, $k = 2$, $\theta_i \sim [\mathcal{U}(3,15)]^5$, $c_{\rm out} = 3$, $c_{\rm in} = 17$. Only nodes $i$ with $d_i > 0$ are considered.\vspace{-0.1cm}}
    \label{fig:my_label}
\end{figure}

We here work on the strong connection between the Bethe-Hessian matrix for $r = \zeta_p$ and the regularized random walk Laplacian. The following equivalent identities indeed hold:
\begin{align*}
[(\zeta_p^2-1)I_n + D - \zeta_pA]\bm{x}_p &\approx 0 \\
D_{\zeta_p^2-1}^{-1}A\bm{x}_p &\approx \frac{1}{\zeta_p}\bm{x}_p,
\label{eq:intuition}
\end{align*}
where $D_{\zeta_p^2-1} = D + (\zeta_p^2-1)I_n$. This notably suggests that, for $\tau = \zeta_p^2-1$, the matrix $L_{\tau}^{\rm rw}$ has an eigenvector whose entries  are not affected by the degree distribution, but depend only on the class labels, as depicted in Figure \ref{fig:my_label}. With Figure \ref{fig:my_label}, consistently with Remark \ref{rem:1}, we further underline that for the \emph{k-means} step it is fundamental to obtain two well separated density clouds in the $k$-dimensional space spanned by the rows of $X$, instead of a continuum of points, as evidenced by the histograms. Since there is a unique value of $r$ that allows the matrix $H_r$ to have a "clean" eigenvector \cite{dall2019revisiting}, also the choice of $\tau$  is unique, as depicted in Figure \ref{fig:ov}. 
An "informative" eigenvector however does not imply that such eigenvector corresponds to a dominant isolated eigenvalue. This is of fundamental importance because, if the informative eigenvector corresponds to a non-isolated eigenvalue, then i) it is algorithmically challenging to locate it and ii) the information is likely to spread out on the neighboring eigenvectors.
This is however not the case thanks to the following two propositions:
\begin{prop}
	\label{prop:1}
	Consider the graph $\mathcal{G}(\mathcal{V},\mathcal{E})$ built on a sparse DC-SBM as per Equation \eqref{eq:DCSBM} with $k$ communities. Further let $\zeta_p$, be defined as in Equation \eqref{eq:zeta}, satisfying $\zeta_p \leq \sqrt{c\Phi}$.
	
	Then, for all large $n$ with high probability,
% the $p$ eigenvalues of $L_{\zeta_p^2-1}^{\rm rw}$ with largest modulus are isolated 
$\zeta_p^{-1}$ is the $p$-th largest eigenvalue of $L_{\zeta_p^2-1}^{\rm rw}$ and it is isolated.
\end{prop}

% We now state the following corollary:

\begin{prop}
	\label{cor:1}
% 	If the $p$ largest eigenvalues of $L_{\tau}^{\rm rw}$ are isolated, then the $p$ largest eigenvalues of $L_{\tau'}^{\rm rw}$ are also isolated for $\tau' > \tau \geq 0$.
The $p$ largest eigenvalues of $L_{\tau}^{\rm rw}$ are isolated, for $\zeta_p^2-1 \leq \tau \leq c\Phi-1$ with high probability for all large $n$.
\end{prop}

Proposition \ref{prop:1} guarantees that, for the proposed parametrization, the informative eigenvector is isolated and can be found in the $p$-th largest position. Thanks to Proposition \ref{cor:1} instead, we know that
% if the regularization is increased, then the uninformative bulk will not spread out. 
for $\tau = c\Phi-1 \geq \zeta_k^2-1$, all the $k$ informative eigenvalues will be isolated.

% \textcolor{blue}{ *** Is it $\geq$ or $>$ ? ***}

We now give a sketch of proof of Proposition \ref{prop:1} and \ref{cor:1}. 

% \textcolor{red}{The extended proofs can be found in the extended version of the article \cite{}.}
% {\nico: the extended results will be presented in \cite{dall2019reg} that is currently under preparation. **this last sentence in red is not necessary. It is ok in conference papers like ICASSP to only show sketches of proofs.**}

\begin{hproof}[Sketch of Proof of Proposition \ref{prop:1}]
Consider the eigenvector equation of $L_{\tau}^{\rm rw}$, for $0 \leq \tau \leq c\Phi-1$,
\begin{equation*}
L_{\tau}^{\rm rw}\bm{x}_p = s_p(L_{\tau}^{\rm rw})\bm{x}_p.
\end{equation*}
We define $r$ by $\tau = r^2-1$, so that $1 \leq r \leq \sqrt{c\Phi}$. The earlier equation can be rewritten in the following form:
\begin{equation*}
rL_{r^2-1}^{\rm rw}\bm{x}_p = s_p(rL_{r^2-1}^{\rm rw})\bm{x}_p.
\end{equation*}
Define $\bar{r}_p$ such that $s_p(\bar{r}_pL_{\bar{r}_p^2-1}) = 1$  (we assume the existence of such an $\bar{r}_p$). Then, necessarily,
\begin{equation*}
H_{\bar{r}_p}\bm{x}_p = 0.
\end{equation*}
From the properties of the Bethe-Hessian matrix, $\bar{r}_p$ can assume only $k+1$ discrete values: $\bar{r}_p \approx \zeta_p~{\rm for}~1\leq p \leq k$ and $\bar{r}_p \approx \sqrt{c\Phi} ~{\rm for}~p = k+1$. Letting $\epsilon \to 0$, then $s_p\left((\bar{r}_p + \epsilon)L_{(\bar{r}_p+\epsilon)^2-1}\right) \neq 1, ~\forall~ 1 \leq p \leq n$, meaning that there is no other eigenvalue in the neighborhood of $s_p(\bar{r}_pL_{\bar{r}_p^2-1}^{\rm rw})$ and hence we conclude it is isolated. As a consequence, the eigenvalues belonging to the \emph{bulk} of $rL_{r^2-1}^{\rm rw}$ are necessarily smaller than $1$ in modulus. 

Intuitively, looking at the symmetry of the spectrum of $L_{\tau}^{\rm rw}$ (Figure \ref{fig:spec}), the isolated eigenvalues are in largest positions. Formally, exploiting the Courant-Fischer theorem one can prove that $1$ is indeed the $p$-th largest eigenvalue of $\zeta_pL_{\zeta_p^2-1}^{\rm rw}$.
We can write:
\begin{subequations}
\begin{align}
s_q^{\downarrow}(rL_{r^2-1}^{\rm rw}) < 1, & \quad \mbox{for }  q \geq  k + 1, \quad r  < \sqrt{c\Phi} \label{eq:pos_bulk} \\
s_p^{\downarrow}(\zeta_pL_{\zeta_p^2-1}^{\rm rw}) = 1, & \quad \mbox{for } p \leq k. \label{eq:pos_info}
\end{align}
\end{subequations}
Equation \eqref{eq:pos_info} states that the informative eigenvector has an eigenvalue equal to $1$, while Equation \eqref{eq:pos_bulk} imposes that all the uninformative eigenvalues belonging to the bulk are smaller than $1$, hence of the informative eigenvalue, so the result.
\end{hproof}

\vspace{-0.5cm}

\begin{hproof}[Sketch of Proof of Proposition \ref{cor:1}]

This proposition is a direct consequence of $s_p^{\uparrow}(D-rA)$ being isolated for $\zeta_p \leq r \leq \sqrt{c\Phi}$ \cite{dall2019revisiting}.

Define $\tilde{r}$ such that, $s_p^{\uparrow}(D-rA) = s_{p+1}^{\uparrow}(D -\tilde{r}A) = -\tau$. By construction, the matrix $L_{\tau}^{\rm rw}$ thus has two eigenvalues that are equal to $r^{-1}$ and ${\tilde{r}}^{-1}$. As a consequence of $s_p^{\uparrow}(D-rA)$ being isolated, $r^{-1}$ is away from $\tilde{r}^{-1}$. 

The change of variable $\tau = - s_p^{\uparrow}(D-rA)$ provides a one-to-one mapping between the smallest isolated eigenvalues of $H_r$ and the largest isolated eigenvalues of $L_{\tau}^{\rm rw}$. It follows that, for $\zeta_p^2-1 \leq \tau \leq c\Phi-1$, the top $p$ eigenvalues of $L_{\tau}^{\rm rw}$ are isolated.

% For $\zeta_p \leq r \leq \sqrt{c\Phi}$, $s_p(r)$  is an isolated eigenvalue and so $s_{p+1}(r) - s_p(r) = O_n(1)$. Given the regularity of $s_p(r)$, we can write
% 	\begin{align*}
% 	O_n(1) = O_n(s_{p+1}(r) - s_p(r)) =  O_n\big(s_{p+1}(r) - s_{p+1}(\tilde{r})\big).
% 	\end{align*}
% 	As a consequence $r^{-1} - \tilde{r}^{-1} = O_n(1)$, so the eigenvalue $1/r$ of $L_{\tau}^{\rm rw}$ is isolated. Since $1/r$ corresponds to the $p$-th largest eigenvalue, the top $p$ eigenvalues of the matrix $L_{\tau}^{\rm rw}$ are isolated.
\end{hproof}

% \begin{hproof}[Sketch of Proof of Corollary \ref{prop:1}]]
% 	For all $p$, the following equation holds
% 	\begin{equation*}
% 	[D - s_p^{\uparrow}(D-rA)I_n]^{-1}A\bm{x}_p = \frac{1}{r}\bm{x}_p.
% 	\end{equation*}
% 	Denoting with $\tau = - s_p^{\uparrow}(D - rA)$, this can be equivalently rewritten as
% 	\begin{equation*}
% 	L_{\tau}^{\rm rw}\bm{x}_p = \frac{1}{r}\bm{x}_p \equiv s^{\tau}_p(r)\bm{x}_p.
% 	\end{equation*}
% 	Consider $\tilde{p}<p$ and define $\tilde{r}$ as $
% 	s_{\tilde{p}}^{\uparrow}(D - \tilde{r}A) = s_p^{\uparrow}(D-rA) = -\tau$ and $\tilde{\tau}$ as $\tilde{\tau} = -s_{\tilde{p}}^{\uparrow}(D - rA) > \tau$.
% 	Then, since $s_p^{\tau}(r)$ is a decreasing function of $\tau$,
% 	\begin{equation*}
% 	s_p^{\tilde{\tau}}(r) < s_p^{\tau}(r) = \frac{1}{r} = s_{\tilde{p}}^{\tilde{\tau}}(r).
% 	\end{equation*}
% 	This allows us to say that, if $s_p^{\tau}(r) < s_{\tilde{p}}^{\tau}(r)$, then for $\tilde{\tau} > \tau$, $s_p^{\tilde{\tau}}(r) < s_{\tilde{p}}^{\tilde{\tau}}(r)$, so the ordering is preserved. 
% \end{hproof}

\vspace{-.5cm}

\subsection{Comments on the result and algorithm.}

\begin{figure}
    \centering
    \includegraphics[width = \columnwidth]{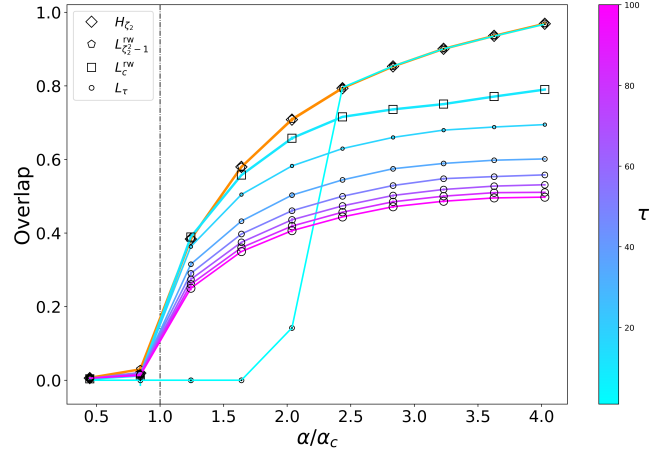}
    \caption{Overlap comparison using the eigenvector with second largest eigenvalue of the matrix $L_{\tau}$ for different values of $\tau$. For these simulations $n = 50.000$ separated in two equal-size classes, $c_{\rm in} = 11 \to 19$, $c_{\rm out} = 9 \to 1$ with $c = 10$ constant, $\theta_i \sim [\mathcal{U}(3,15)]^5$. Averages are taken over $5$ realizations. The color code indicates the values of $\tau = 1 \to c^2$ also encoded by the size of the dots (small dots are for small $\tau$). The orange line with pentagons is obtained for $L_{\zeta_2^2-1}^{\rm rw}$, the line with squares is obtained for $L_{c}^{\rm rw}$. The orange line with diamonds is for $H_{\zeta_2}$.}
    \label{fig:ov}
\end{figure}

Figure~\ref{fig:ov} compares the performance of reconstruction in terms of the overlap
\begin{align}
    Ov = \left(\frac{1}{n}\sum_i \delta(\hat{\ell}_i,\ell_i) - \frac{1}{k}\right)\frac{1}{1-\frac{1}{k}}
    \label{eq:ov}
\end{align}
and evidences that i) small regularizations produce better node partitions, ii) the proposed, $\alpha$-dependent, regularization surpasses all fixed values of $\tau$, iii) for $\tau < \zeta_2^2-1$ good partitions are achieved on easy problems, but the information does not correspond to isolated eigenvectors for hard detection problems, and iv) the performance using $H_{\zeta_2}$ and $L_{\zeta_2^2-1}^{\rm rw}$ are the same, since they are using the same informative eigenvectors.

We next list important messages of Proposition~\ref{prop:1} and \ref{cor:1}.

\begin{algorithm}[b!]
\normalsize
	\begin{algorithmic}[1]
		\State \textbf{Input} : adjacency matrix of undirected graph $\mathcal{G}$
		\State Estimate $k$~:~  $\hat{k} \leftarrow \left|\left\{i,~ s_i(D_{c\Phi-1}^{-1}A) > \frac{1}{\sqrt{c\Phi}} \right\}\right|$.
		\For{$1 \leq p \leq \hat{k}$}
		\State $\zeta_p \leftarrow\underset{r}{\rm arg}\left[s_p^{\uparrow}(H_r) = 0\right] $
		\State $X_p \leftarrow \bm{x}_p : \zeta_pD_{\zeta_p^2-1}^{-1}A\bm{x}_p =  \bm{x}_p $
		\EndFor
		\State Estimate community labels $\hat{\bm{\ell}}$ as output of $\hat{k}$-class \emph{k-means} on the rows of $X = [X_2, \ldots, X_{\hat{k}}]$.\\
		\Return Estimated number $\hat{k}$ of communities and label vector $\hat{\bm{\ell}}$.
		\caption{Community Detection with the regularized Laplacian}
		\label{alg:2}
	\end{algorithmic}
\end{algorithm}

\begin{enumerate}[leftmargin=0.1in]
	\setlength\itemsep{0em}
	\item \emph{$L^{\rm sym}_{\tau}$ vs. $L_{\tau}^{\rm rw}$} : as opposed to \cite{qin2013regularized}, we studied the matrix $L_{\tau}^{\rm rw}$ instead of $L^{\rm sym}_{\tau}$. These two matrices have the same eigenvalues, but not the same eigenvectors. Our line of argument suggests -- in good agreement with the observation of \cite{von2007tutorial} -- that it is more convenient to use the eigenvectors of the matrix $L_{\tau}^{\rm rw}$ in order to obtain eigenvectors whose entries are not affected by the degree distribution. 
	\item \emph{The value of $\tau$ at the transition}: consider $k = 2$, then $\zeta_2 = (c_{\rm in} + c_{\rm out})/(c_{\rm in}  - c_{\rm out})$. When $\alpha = \alpha_c$ (Equation \eqref{eq:th}), $\zeta_2 = \sqrt{c\Phi}$ and therefore $\tau = c\Phi - 1 \approx c$. This observation allows us to understand why, in practice, the regularization $\tau = c$ appears to be a good choice. When $\tau = c\Phi - 1$ then certainly -- regardless the hardness of the problem, as long as $\alpha > \alpha_c$ -- the second largest eigenvalue of $L_{\tau}^{\rm rw}$ is isolated. As $c$ is in the order of magnitude of $c\Phi - 1$ in sparse graphs, $\tau = c$ will lead -- in most cases -- to a "satisfying" spectrum, in the sense that the informative eigenvalues are isolated. When $k>2$, $\zeta_p = \sqrt{c\Phi}$ represents the transition from easy to hard detection and the argument can be generalized.
	\item \emph{The regularization is a function of the hardness of the detection problem}: once again consider $k = 2$. For easy detection problems ($c_{\rm out}\to 0$), we have $\zeta_2 = (c_{\rm in} + c_{\rm out})/(c_{\rm in} - c_{\rm out}) \to 1$, while, as already mentioned in point 2, $\zeta_2$ increases up to $\sqrt{c\Phi}$ in harder scenarios. This implies that harder problems need a larger regularization. Note also that, in the trivial case for which $\zeta_p \to 1,~\forall~ p$, (when we have $k$ nearly disconnected clusters), the Bethe-Hessian falls into the combinatorial graph Laplacian $\lim_{r \to 1} H_r = D - A$ and the regularized random walk Laplacian into its non-regularized counterpart $\lim_{\tau \to 0}L_{\tau}^{\rm rw} = L^{\rm rw}$.
	\item \emph{Estimating the values of $\zeta_p$}: thanks to Equation \eqref{eq:zeta} the values of $\zeta_p$ can be obtained by searching for the solution to $s_p^{\uparrow}(H_r) = 0$ on $r \in (1,\sqrt{c\Phi})$. 
% 	Alternatively, they can be obtained from the ratio of the largest eigenvalues of the matrix $B'$ whose leading eigenvalues are proved to converge in the large $n$ limit to $s_p^{\downarrow}(B') = s_p^{\downarrow}(C\Pi)\Phi$ \cite{gulikers2016non}:
% 	\begin{align}
% 	B' = 
% 	\begin{pmatrix} 
% 	A & -I_n \\
% 	D - I_n & 0
% 	\end{pmatrix}.
% 	\end{align}
	\item \emph{Estimating the number of classes}: it was shown in \cite{saade2014spectral} that all and only the informative eigenvalues of $H_r$ at $r = \sqrt{c\Phi}$ are negative, allowing an unsupervised method to estimate $k$. From this result, also $L_{\tau}^{\rm rw}$ allows to estimate $k$ as follows:
	\begin{align}
	\hat{k} = \left|\{i : s_i(L^{\rm rw}_{c\Phi-1}) > \frac{1}{\sqrt{c\Phi}}\}\right|
	\end{align}
	\item \emph{Disassortative networks}: even though we assumed for simplicity all the eigenvalues of the matrix $C\Pi$ to be positive -- hence that there is a larger probability to get connected to nodes in the same community (\emph{assortativity}) -- the above results can be easily generalized to the case in which $C\Pi$ has negative eigenvalues and so in which there are \emph{disassortative} communities.
\end{enumerate}

The results of Section \ref{sec:main.proof} naturally unfold in Algorithm \ref{alg:2} for community detection in sparse graphs.

\section{Conclusion}
\label{sec:conclusion}

In this article we discussed the regularization parameter $\tau$ of the matrix $L_{\tau}^{\rm rw} = (D + \tau I_n)^{-1}A$, used to reconstruct communities in sparse graphs. We explained why small and, most importantly, difficulty-adapted regularizations perform better than large (and difficulty-agnostic)  ones.

Our findings notably shed light on the connection between the two benchmark approaches to community detection in sparse networks, provided for one by the statistics community and for the other by the physics community; these approaches have so far have been treated independently. We strongly suggest that bridging both sets of results has the capability to improve state-of-the-art knowledge of machine learning algorithms in sparse conditions (for which a direct application of standard algorithms is often inappropriate). Similar outcomes could arise for instance in KNN-based kernel learning or for any algorithm involving numerous data which, for computational reasons, imposes a sparsification of the information matrices.

In this view, we notably intend to generalize the algorithm proposed in this article (i) to richer graph and data clustering problems, (ii) to the often more realistic semi-supervised setting (where part of the nodes are known), while in passing (iii) enriching our understanding on existing algorithms.

\section{Acknowledgements}

Couillet's work is supported by the IDEX GSTATS DataScience Chair and the MIAI LargeDATA Chair at University Grenoble Alpes. Tremblay's work is supported by CNRS PEPS I3A (Project RW4SPEC).

% References should be produced using the bibtex program from suitable
% BiBTeX files (here: strings, refs, manuals). The IEEEbib.bst bibliography
% style file from IEEE produces unsorted bibliography list.
% -------------------------------------------------------------------------

\end{document}